\documentclass[preprint,12pt]{elsarticle}
\usepackage{amsmath}
\usepackage[]{algorithm2e}

\usepackage{ulem}
\usepackage{amssymb}

\journal{Information Sciences}

\begin{document}

\begin{frontmatter}



\title{Fast Image Classification by Boosting Fuzzy Classifiers}

\ead[url]{http://iisi.pcz.pl/}
\author{Marcin Korytkowski}
\ead{marcin.korytkowski@iisi.pcz.pl}
\author{Leszek Rutkowski}
\ead{leszek.rutkowski@iisi.pcz.pl}
\author{Rafa{\l} Scherer}
\ead{rafal.scherer@iisi.pcz.pl}

\address{Institute of Computational Intelligence, Cz\c{e}stochowa University of Technology, al. Armii Krajowej 36, 42-200 Cz\c{e}stochowa, Poland}
\begin{abstract}
This paper presents a novel approach to visual objects classification based on generating simple fuzzy classifiers using local image features to distinguish between one known class and other classes. Boosting meta learning is used to find the most representative local features. The proposed approach is tested on a state-of-the-art image dataset and compared with the bag-of-features image representation model combined with the Support Vector Machine classification. The novel method gives better classification accuracy and the time of learning and testing  process is more than 30\% shorter.
\end{abstract}


\end{frontmatter}


\section{Introduction}
\label{sec:intro}
The most popular way to search vast collections of images and video which are generated every day in a tremendous amount is realized by keywords and meta tags or just by  browsing them. Emergence of content-based image retrieval (CBIR) in the 1990s enabled automatic retrieval of images to a certain extent. Various CBIR tasks include searching for images similar  to the query image or retrieving images of a certain class \cite{GuimaraesPedronette201491}\cite{DrozdaSG13}\cite{Kanimozhi20151099}\cite{Karakasis201522}\cite{Lin20146611}\cite{Liu2013188}\cite{Liu2012744}\cite{Rashedi201385}\cite{Wang201543}\cite{Wu20132927} \cite{Yu2013355}  and classification \cite{7083681}\citep{JAISCR-optimized}\citep{JAISCR-improved}\cite{jegou2010aggregating}\cite{jegou2012aggregating}\cite{JAISCR-breast}\cite{Liu2014}\cite{Shrivastava2014212}\cite{5206757_CVPR2009}  of the query image. Such content-based image matching remains  a challenging problem of computer science. Image matching consists of two relatively difficult tasks: identifying objects on images and fast searching through large collections of identified objects. 
Identifying objects on images is still a challenge as the same objects and scenes can be viewed under different imaging conditions. There are many previous works dedicated to the problem formulated in this way. Some of them are based on color representation \cite{Huang1997correlograms}\cite{Kiranyaz:2010}\cite{PassZabih1996}, textures \cite{ChangTexture1993}\cite{FrancosUnifiedTexture1993}\cite{JainGabor1991}\cite{smietanski2010texture}, shape \cite{Jagadish:1991}\cite{Kauppinen1995}\cite{Veltkamp:2000Shape} or edge detectors \cite{Zitnick:2014}\cite{ogiela2002syntactic}\cite{ogiela2005nonlinear}. Recently local invariant features have gained a wide popularity \cite{SIFT:Lowe:2004}\cite{Matas2004761}\cite{Mikolajczyk2004}\cite{Nister:2006}\cite{SivicVideoGoogle}. 
The most popular local keypoint detectors and descriptors are SURF \cite{SURF:Bay:2008}, SIFT \cite{SIFT:Lowe:2004} or ORB \cite{Rublee2011}.

The second  substantial problem is a fast retrieval of identified objects.  
To find similar images to a query image, we need to compare all feature descriptors of all images usually by some distance measures. Such comparison is enormously time consuming and there is ongoing worldwide research to speed up the process. Yet, the current state of the art in the case of high dimensional computer vision applications is not fully satisfactory. The literature presents  countless methods and variants utilizing e.g. a voting scheme or histograms of clustered keypoints. 
They are mostly based on some form of approximate search. 
One of the solutions to the problem can be  descriptor vector hashing.  In \cite{Datar:2004:LHS} the authors proposed a locality-sensitive hashing method for the approximate nearest neighbour algorithm.  In \cite{Nister:2006} the authors built a hierarchical quantizer in the form of a tree. Such tree is a kind of an approximate nearest neighbour algorithm and constitutes a visual dictionary. 
Recently, the bag-of-features (BoF)  approach \cite{Grauman2005}\cite{Philbin2007}\cite{SivicVideoGoogle}\cite{SlavaBoF}\cite{ZhangMarszalek2006} has  gained in popularity. In the BoF method clustered vectors of image features are collected and sorted by the count of occurrence (histograms). There are some modifications of this method, for example a solution that uses earth mover’s distance (EMD) presented in \cite{Grauman2005}. The main problem with the aforementioned approaches is that all individual descriptors or approximations of sets of descriptors presented in the histogram form must be compared. Such calculations are very computationally expensive. Moreover, the BoF approach requires redesign of the classifier when new visual classes are added to the system. In this paper we will overcome these limitations and we will present a novel method to  classify fast various images in large collections on the basis of their content. 
Our method is partly inspired by the ideas of Viola et al. \cite{Tieu2004}\cite{Viola2001}\cite{Zhang2005}. They used a modified version of the AdaBoost algorithm to select the most important features from a large number of very simple rectangular features similar to Haar basis functions. The selected features are treated by the authors of \cite{Tieu2004}\cite{Viola2001} as weak classifiers for the content-based image retrieval task, mainly   images containing faces. Contrary to the previous authors, who developed CBIR systems based on boosting techniques, in the approach proposed in this paper: (i) we use the original version AdaBoost algorithm to choose the most important local features; (ii) a wide variety of local and global visual features descriptors (e.g. SURF SIFT or ERB) can be incorporated into our classifier; (iii) our method is applicable  to a wider class of images (not only face images); (iv) incorporating new visual classes in the system requires only adding new fuzzy rules to the rule base without restructuring the existing rule base.     
In this paper we propose a novel approach to use fuzzy logic and fuzzy rules as the adjustable representation of visual feature clusters. Fuzzy logic \cite{rutkowska2000new}\cite{RutkL:Kluwer:2004}\cite{RutkL:Springer:2008}\cite{rafsch2010IJNS}\cite{scherer2014multiple} is a very convenient method for describing partial membership to a set. This allows creating a very efficient method for fast object classification in large databases.  We combine fuzzy logic and boosting meta learning to choose the most representative set of image features for every considered class of objects. In each step we randomly choose one feature from a set of positive images taking into consideration feature weights computed using the Adaboost algorithm.  This feature constitutes a base to build a weak classifier. The weak classifier is given in the form of a fuzzy rule and the selected feature is a base to determine the initial parameters of the fuzzy rule. In the learning process the weak classifiers are adjusted to fit positive image examples. This approach could be very useful for the search based on the image content in a set of complex graphical objects in a database.
The main contribution and novelty of the paper is as follows:
\begin{itemize}
\item  We present a novel method for automatic building a fuzzy rule base for image classification based on local features; the method does not require knowledge of any initial parameters, contrary to e.g.  the popular BoF method which requires specifying the dictionary size.
\item We develop an efficient technique for fast classification of images, in particular the learning  and the testing time of our method is, respectively, 35\% and 32\% shorter than in the case of BoF.
\item We propose a method for automatic search of the most salient local features for a given class of images.
\item We design a flexible system; expanding the system knowledge is efficient because adding new visual classes to the system requires only adding new fuzzy rules whereas in the case of BoF it requires the whole new dictionary generation and re-learning of classifiers.
\end{itemize}

The paper is organized as follows. Section \ref{sec:learning} describes the proposed method of creating the weak classifier ensemble as a fuzzy rule base. The method of classifying new query images is presented in Section  \ref{sec:testing}. Section \ref{sec:experiments} provides simulation results on the the PASCAL Visual Object Classes (VOC) 2012 dataset \cite{Everingham10} to compare the proposed method with the BoF method combined with support vector machines classifiers. 
\section{Boosting-Generated Simple Fuzzy Classifiers}
\label{sec:learning}
The main idea of this paper is to find the most representative fuzzy rules for a given class $\omega_c$, $c=1,\ldots,V$, of visual objects and to fast classify query images afterwards. This section describes the learning process i.e. generating fuzzy rules from a set of examples. The algorithm uses the boosting meta learning to generate suitable number of weak classifiers.
%
%
%
%
%
The classifiers feature space $\mathbb{R}^N$ consists of elements $x_n$, $n=1,..,N$. For example, in the case of using the standard SIFT descriptors, $N=128$. 

 In each step we randomly choose one local feature from the set of positive images according to its boosting weight. Then we search for similar feature vectors from all positive images. Using these similar features we construct one fuzzy rule. 
 Undoubtedly, it is impossible to find exactly the same features in all images from the same class, thus we search for feature vectors which are similar to the feature picked randomly in the current step.  This is the one of the reasons for using fuzzy sets and fuzzy logic.  
The rules have the following form:
\begin{equation}
\label{eq:Rules}
\begin{array}{c}
R^{c}_t\colon\text{IF }x_{1}\text{ is }G_{1,t}^{c} \text{ AND }x_{2} \text{ is }G_{2,t}^{c}\text{ AND}\ldots\\
\ldots\text{ AND }x_{128}\text{ is }G_{128,t}^{c} \text{ THEN }
\text{image } i \in\ \omega_c (\beta_{t}^c) 
\end{array}
\text{,}
\end{equation}
where $t=1,\ldots,T^c$ is the rule number in the current run of boosting, $T^c$ is the number of rules for the class $\omega_c$
and $\beta_{t}^c$ is the importance of the classifier, designed to classify objects from the class $\omega_c$, created in the $t$-th boosting run. 
 The weak classifiers \eqref{eq:Rules} consist of fuzzy sets with Gaussian membership functions
\begin{equation}
\label{eq:Gaussian}
G_{n,t}^{c}(x)=\mathop{e}\nolimits^{-\left(\frac{x-m_{n,t}^{c}}{\sigma_{n,t}^{c}} \right)^{2} } ,
\end{equation}
where $m_{n,t}^{c}$ is the center of the Gaussian function \eqref{eq:Gaussian} and $\sigma_{n,t}^{c}$ is its width. 
For the clarity of presentation this section describes generating the ensemble of weak classifiers for a class $\omega_c$, thus the class index $c$ will be omitted. 
\
Let $I$ be the number of all images in the learning set, divided into two sets: positive images  and   negative images, having, respectively, $I_{pos}$ and $I_{neg}$ elements. Obviously $I=I_{pos}+I_{neg}$. Positive images belong to a class $\omega_c$ that we train our classifier with. For every image from these two sets we determine local features, for example local interest points using e.g. SIFT or SURF algorithms. 
The points are represented by descriptors, and we operate on two sets of vectors:  positive descriptors $\{\mathbf{p}^i;i=1,..,L_{pos}\}$ and negative ones $\{\mathbf{n}^j;j=1,..,L_{neg}\}$. In the case of the standard SIFT algorithm, each vector ${{\mathbf p}}^i$ and ${{\mathbf n}}^j$  consists of 128 real values. Let ${v^i}$ be the number of keypoint vectors in the $i$th positive image, let ${u^j}$ be the number of keypoint vectors in the $j$th negative image. Then, the total number of learning vectors is given by 
\begin{equation}
\label{eq:sumaposineg}
L=\sum\limits_{i = 1}^{I_{pos}} {v^i} +  \sum\limits_{j = 1}^{I_{neg}} {u^j} \;,
\end{equation}
where $L=L_{pos}+L_{neg}$.
According to the AdaBoost algorithm, we have to assign a weight to each keypoint in the learning set. When creating new classifiers the weights are used  to indicate keypoints which were difficult to handle. At the start of the algorithm, all the weights have the same, normalized values
\begin{equation}
D^l_1=\frac{1}{L} \text{ for } l=1,\ldots,L \;.
\end{equation}
Let us define matrices $\textbf{P}_t$ and $\textbf{N}_t$ constituting the learning set
\begin{equation}
\label{matrix_P_t}
{\mathbf P_t}=\left[ 
\begin{matrix}
{{\mathbf p}}^{1} & D^1_t \\ 
 \vdots    & \vdots  \\
{{\mathbf p}}^{L_{pos}} & D_t^{L_{pos}} \end{matrix}
\right]=\left[ \begin{matrix}
p^{1}_1,\dots ,p^{1}_{N} & D^1_t \\ 
 \vdots    & \vdots  \\
p_{1}^{L_{pos}},\dots ,p_{N}^{L_{pos}} & D_t^{L_{pos}} \end{matrix}
\right] \;,
\end{equation}
\begin{equation}
{\mathbf N_t}=\left[ \begin{matrix}
{{\mathbf n}}^{1} & D^1 \\ 
 \vdots    & \vdots  \\
{{\mathbf n}}^{L_{neg}} & D_t^{L_{neg}} \end{matrix}
\right]=\left[ \begin{matrix}
n^{1}_{1},\dots ,p_{N}^{1} & D^1 \\ 
 \vdots    & \vdots  \\
n_{1}^{L_{neg}},\dots ,p^{L_{neg}}_{N} & D_t^{L_{neg}} \end{matrix}
\right] \;.
\end{equation}
The learning process consists in creating $T$ simple classifiers (weak learners in boosting terminology) in the form of fuzzy rules \eqref{eq:Rules}. After each run $t$, $t=1,\ldots,T$, of the proposed algorithm, one fuzzy rule $R_t$ is obtained. The process of building a single fuzzy classifier is presented below. 
\begin{enumerate}
\item  Randomly choose one vector ${{\mathbf p}}^r,\ 1\le r\le L_{pos}$  from positive samples using normalized distribution of elements $D^1_t,\ldots, D_t^{L_{pos}}$ in matrix \eqref{matrix_P_t}.
This drawn vector becomes a basis to generate a new classifier and the learning set weights contribute to the probability of choosing a keypoint.

\item  For each image from the positive image set find the feature vector which is nearest to ${{\mathbf p}}^r$ (for example according to the Euclidean distance) and store this vector in matrix ${\mathbf M}_t$ of the size
 $I_p \times N$. Every row represents one feature from a different image $v_i$, $i=1,\ldots,I_{pos}$, and no image occurs more than once
\begin{equation}
\label{eq:macierzMt}
\mathbf{M}_t=\left[
\begin{matrix} 
{\tilde{p}_{t,1}^{1} } & \cdots & \tilde{p}_{t,N}^{1} \\ 
 \vdots    & \cdots & \vdots  \\
{\tilde{p}_{t,1}^{j} } & \ddots & \tilde{j}_{t,N}^{j} \\ 
 \vdots    & \cdots & \vdots  \\
{\tilde{p}_{t,1}^{I_{pos}} } & \cdots & \tilde{p}_{t,N}^{I_{pos}} 
\end{matrix}
\right], 
\end{equation} 
Each vector $\begin{bmatrix}
{\tilde{p}_{t,1}^{j} } & \cdots & \tilde{p}_{t,N}^{j} 
\end{bmatrix}$, $j=1,\ldots,I_{pos}$, in matrix \eqref{eq:macierzMt} contains one visual  descriptor from the set $\{\mathbf{p}^i;i=1,..,L_{pos}\}$. For example, in view of descriptions \eqref{matrix_P_t} and \eqref{eq:sumaposineg}, the first row in matrix \eqref{eq:macierzMt} is one of the rows of the following matrix\\
\begin{equation}
\label{eq:macierzkawalekPt}
\begin{bmatrix}
p^{1}_1,\dots ,p^{1}_{N} \\ 
 \vdots    & \vdots  \\
p_{1}^{v_{1}},\dots ,p_{N}^{v_{1}} \end{bmatrix}
 \;,
\end{equation} 
where $v_1$ is the number of feature vectors in the first positive image.
%
%
%
\item  In this step a weak classifier is built, i.e. we find centres and widths of Gaussian functions which are membership functions of fuzzy sets in a fuzzy rule \eqref{eq:Rules}. 
\begin{enumerate}


\item  Compute the absolute value  $d_{t,n}$ as the difference between the smallest and the highest values in each column of the matrix \eqref{eq:macierzMt}
\begin{equation}
d_{t,n}=\lvert \min_{i=1,\ldots,I_p} p_n^i - \max_{i=1,\ldots,I_p} p_n^i \rvert
\end{equation}
where $n=1,\ldots,N$. Compute the center of fuzzy Gaussian  membership function (\ref{eq:Gaussian}) $m_{t,n}$ in the following way
\begin{equation}
m_{t,n}={\max_{i=1,\ldots,I_p}}{p_n^i} - \frac{d_{t,n}}{2} \;.
\end{equation}
Now we have to find the widths of these fuzzy set membership functions. We have to assume that for all real arguments in the range of $\left[m_{t,n}-\frac{d_{t,n}}{2};m_{t,n}+\frac{d_{t,n}}{2}\right] $ the Gaussian function (fuzzy set membership function) values should satisfy $G_{n,t}(x) \geq 0.5$. Only in this situation do we activate the fuzzy rule. As we assume that $G_{n,t}(x)$ is at least 0.5 to activate a fuzzy rule, using simple substitution $x=m_{t,n}-\frac{d_{t,n}}{2} $, we obtain the relationship for $\sigma_{t,n}$
\begin{equation}
 \sigma_{t,n}=\frac{d_{t,n}}{2\sqrt{-\ln (0.5)} } 
\end{equation} 

Finally, we have to calculate the values $m_{t,n}$ and $\sigma_{n,t}$ for every element of the $n$th column of matrix \eqref{eq:macierzMt}, thus we have to repeat the above steps for all $N$ dimensions. In this way, we obtain $N$ Gaussian membership functions of $N$ fuzzy sets. Of course, we can label them using fuzzy linguistic expressions such as 'small', 'large' etc., but for the time being we mark them only in a mathematical sense by $G_{n,t}$, where  
$n$, $n=1,..,N$, is the index associated with feature vector elements  and $t$ means the fuzzy rule number. 
\item   Using values obtained in point a) we can construct a fuzzy rule which creates a fuzzy classifier \eqref{eq:Rules}.
\end{enumerate}
\item  Now we have to evaluate the quality of the classifier obtained in step 3. We do this using the standard AdaBoost algorithm \cite{schapire99}. Let us determine the activation level of the rule $R_t$ which is computed by a t-norm of all fuzzy sets membership function values 
\begin{equation}
\label{eq:activation_level_Tnorm}
f_{t}(\bar{\mathbf{x}})= \mathop{\mathop{T}\limits^{N} }\limits_{n=1}G_{n,t}(\bar{x}_n) \;,
\end{equation}
where $\bar{\mathbf{x}}= \left[\overline{x}_1,\ldots,\overline{x}_N \right]$ is a vector of values of linguistic variables $x_1,\ldots,x_N$. 
In the case of minimum t-norm formula \eqref{eq:activation_level_Tnorm} becomes
\begin{equation}
\label{eq:activation_level_minim}
f_{t}(\bar{\mathbf{x}})= \mathop{\mathop{\min}\limits^{N} }\limits_{n=1}G_{n,t}(\overline{x}_n) \;.
\end{equation}
As a current run of the AdaBoost is for a given class $\omega_c$, we can treat the problem as a binary classification (dichotomy)
i.e.  
$y^l=1$ for descriptors of positive images and $y^l=0$ for descriptors of negative images. Then the fuzzy classifier decision  is computed by 
\begin{equation}
\label{eq:ht_od_x}
h_{t}(\bar{\mathbf{x}}^l)= \left\{ {\begin{array}{*{20}l}
   {\begin{array}{*{20}l}
   1 & {{\rm{if}}} & f_{t}(\bar{\mathbf{x}}^l)\geq \frac{1}{2}  \\
\end{array}}  \\
   {\begin{array}{*{20}c}
   0 & {{\rm{otherwise}}} &   \\
\end{array}}  \\
\end{array}} \right.
\;.
\end{equation}
For all the keypoints stored in matrices $\textbf{P}_t$ and $\textbf{N}_t$ we calculate new weights $D_t^l $. To this end, we compute the error of classifier \eqref{eq:ht_od_x} for all $L=L_{pos}+L_{neg}$ descriptors of all positive  and negative images
%
\begin{equation}
\label{algG_boost}
\varepsilon _t  = \sum\limits_{l = 1}^L {D_t^l I(h_t (\bar{\mathbf{x}}^l) \ne y^l )} 
\; ,\end{equation}
where $I$ is the indicator function 
\begin{equation}
I(a \ne b) = \left\{ {\begin{array}{*{20}c}
   {\begin{array}{*{20}c}
   1 & {{\rm{if}}} & {a \ne b}  \\
\end{array}}  \\
   {\begin{array}{*{20}c}
   0 & {{\rm{if}}} & {a = b}  \\
\end{array}}  \\
\end{array}} \right.
\; .\end{equation}
If $\varepsilon _t =0$ or $\varepsilon _t > 0.5$, we finish the training stage. If not, we compute new weights:
\begin{equation}
\label{eq:alfa_inlearning}
\alpha _t  = 0.5\ln \frac{{1 - \varepsilon _t }}{{\varepsilon _t }} \;.
\end{equation}
\begin{equation}
\label{eq:wagiBoosting_inlearning}
D_{t + 1}^l = \frac{{D_t^l\exp \{  - \alpha _t I(h_t(\bar{\mathbf{x}}^l) = y^l )\} }}{{C }}
\; ,\end{equation}
where $C$ is a constant such that
$
\sum_{l = 1}^L {D_{t + 1}^l  = 1} 
\; .$ Finally classifier importance is determined by
\begin{equation}
\label{eq:Boosting_importance_inlearning}
\beta_t=\frac{\alpha_t}{\sum_{t=1}^T\alpha_t}
\; .
\end{equation}
\end{enumerate}
\subsubsection*{Remark 1}
It should be noted that the classifier importance \eqref{eq:Boosting_importance_inlearning} is needed to compute overall response of the boosting ensemble for the query image, which will be described in detail in the next section. 
\subsubsection*{Remark 2}
The concept of 'word' used in the BoW method \cite{Grauman2005}\cite{Philbin2007}\cite{SivicVideoGoogle}\cite{ZhangMarszalek2006} corresponds to a fuzzy rule in the presented method, which in the case of the SIFT application, consists of 128 Gaussian functions. 
\vspace{0.5cm}

The next section will describe a classification of a new query image by the ensemble. 
%
%
%
%
%
%
%
%
%
%
%
%
%
%
%
%
%
%
%
%
%
%
%
%
%
\section{Classification of a Query Image}
\label{sec:testing}
The boosting procedure described in the previous section should be executed for every visual object class $\omega_c$, $c=1,\ldots,V$, thus after the learning procedure we obtain a set of $V$ strong classifiers.
Let us assume that we have a new query image and an associated set of \textit{u} visual features represented by matrix  $\textbf{Q}$ 
\begin{equation}  
\textbf{Q}=\begin{bmatrix}{\textbf{q}^{1} } \\ {\textbf{q}^{2} } \\ {\vdots } \\ {\textbf{q}^{u} } \end{bmatrix}=
\begin{bmatrix}
{q_{1}^{1} \dots q_{N}^{1} } \\ {q_{1}^{2} \cdots q_{N}^{2} } \\ {\vdots } \\ {q_{1}^{u} \cdots q_{N}^{u} } 
\end{bmatrix}\;.
\end{equation} 
Let us determine the value of  
\begin{equation}  
\label{eq:adaboost_weak_out_multi}
F_{t}(\textbf{Q})=\mathop{\mathop{S}\limits^{u}}\limits_{j=1}\left(
\mathop{\mathop{T}\limits^{N}}\limits_{n=1}G_{n,t} (q_{n}^{j} )\right)
%
\; ,
\end{equation}  
%
%
where $S$ and $T$ are $t$-norm and $t$-conorm, respectively (see \citep{RutkL:Springer:2008}). To compute the overall output of the ensemble of classifiers designed in Section \ref{sec:learning}, for each class $\omega_c$ we sum weak classifiers outputs \eqref{eq:adaboost_weak_out_multi} taking into consideration their importance  \eqref{eq:Boosting_importance_inlearning}, i.e. 
\begin{equation} 
\label{eq:adaboost_strong_out_multi} 
H^{c}(\textbf{Q}) =\sum\limits_{t = 1}^{T^c} {\beta_t F_t (\textbf{Q})}  \;.
\end{equation}
Eventually, we assign a class label to the query image in the following way
\begin{equation}  
\label{eq:adaboost_finaldecision_multi}
f(\textbf{Q}) =\arg  \mathop{\max_{c=1,\ldots,V}}  H^{*c}(\textbf{Q})\;.
\end{equation}
In formulas \eqref{eq:adaboost_strong_out_multi}  and \eqref{eq:adaboost_finaldecision_multi} we restored class label index $c$, which had been removed at the beginning of Section \ref{sec:learning}. In formula \eqref{eq:adaboost_weak_out_multi} $t$-norm and $t$-conorm can be chosen as min and max operators, respectively.
\section{Experiments}
\label{sec:experiments}
\begin{figure}
\centering
\includegraphics[width=13cm]{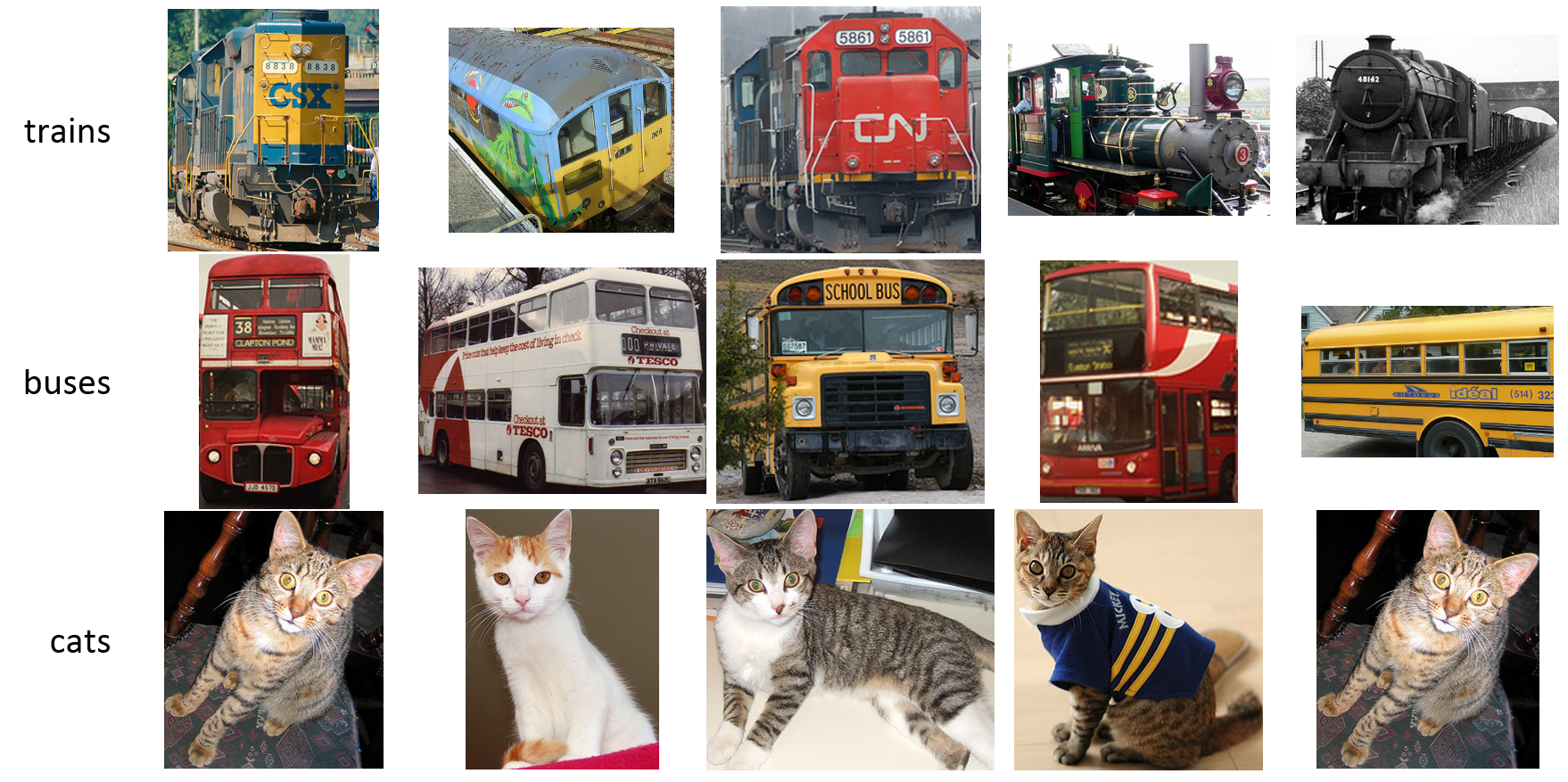}
\caption{Examples of objects from Bus, Cat and Train class taken from PASCAL Visual Object Classes (VOC) dataset.}
\label{fig:imagesVOC}       
\end{figure}

The goal of the experiments was to evaluate the proposed approach and compare it with the state-of-the-art method in terms of accuracy and speed. 
We tested the proposed method on three classes of visual objects taken from the PASCAL Visual Object Classes (VOC) dataset \cite{Everingham10}, namely: Bus, Cat and Train. Examples of such visual objects are presented in Fig. \ref{fig:imagesVOC}. We divided these three classes of objects into learning and testing examples. The testing set consists of 15\% images from the whole dataset. Before the learning procedure we generated local keypoint vectors for all images from the Pascal VOC dataset using the SIFT algorithm. These 128-element vectors were stored in separate files for every image in the dataset. Each file contained hundreds of vectors, depending on the complexity of the image. 

All simulations were performed on a Hyper-V virtual machine with MS Windows Operating System (8 GB RAM, Intel Xeon X5650, 2.67 GHz). We determined the time and quality of classification. The testing set contained only images that had never been presented to the system during learning process.

The method proposed in the paper was implemented using C\# language. For the learning process needs, we built an extra set of examples (negative examples) for each considered class of objects. The negative examples were picked by random from other classes. We chose only the most representative objects for considered classes, whereas for the testing purposes we chose various kinds of images from a considered class. 

We have compared our results with the bag-of-features image representation model combined with the Support Vector Machine (SVM) classification based on the Chi-Square kernel. The BoF algorithm is currently one of the most popular algorithms in computer vision and it was run five times for various dictionary sizes: 200, 250, 300, 350 and 400 words. Dictionaries for the BoF were created using C++ language, based on the OpenCV Library \cite{bradski2000opencv}. 
The BoF experiments were performed on the same set of objects as the experiments for the  method proposed in the paper. 
The results of the BoF and SVM classification, both learning and testing,  are presented in Table \ref{tab:BOF}. 
%
%
%
%
%
%
%
%
%
%
%
%
\begin{table}
\caption{Results of the learning and testing processes for the BoF and SVM algorithms (CQ - Classification Quality ([\%]), LT - Learning time ([s]), TT - Testing time ([s])). The learning time is given only as the overall time for all classes.} 
\label{tab:BOF}
\centering
\footnotesize
\begin{tabular}{|p{0.5in}|p{0.4in}|p{0.5in}|p{0.6in}|} \hline 
~ & \multicolumn{3}{|c|}{Dictionary size: 200} \\ \hline 
~ & CQ & LT  & TT  \\ \hline 
Buses & 70.59\% & ~ & 3.532 	\\ \hline 
Cats & 100\% & ~ & 5.199 		\\ \hline 
Trains & 41.17\% & ~ &	4.833  \\ \hline 
Total & 41.18\% & 195.57 & 13.564 	 \\ \hline
~ & \multicolumn{3}{|c|}{Dictionary size: 250} \\ \hline 
~ &  CQ & LT  & TT  \\ \hline 
Buses & 70.59\% 	& ~ 	& 3.627 \\ \hline 
Cats & 88.24\% 	& ~ 	& 5.858 \\ \hline 
Trains & 35.29\% 	& ~ 	& 5.177 \\ \hline 
Total  	& 64.71\% 	& 208.241	 & 14.662 \\ \hline
~ & \multicolumn{3}{|c|}{Dictionary size 300}\\ \hline 
~ & CQ & LT  & TT    \\ \hline 
Buses & 76.47\% & ~ & 3.678 	\\ \hline 
Cats & 88.24\% & ~ & 5.734 		\\ \hline 
Trains & 41.18\% & ~ & 5.134	 \\ \hline 
Total & 68.63\% & 213.317 & 14.546 	 \\ \hline 
~ & \multicolumn{3}{|c|}{Dictionary size 350} \\ 
\hline 
~ & CQ & LT  & TT  \\ \hline 
Buses & 70.59\% & ~ 			& 3.696 \\ \hline 
Cats & 94.12\% & ~ 			& 5.862 \\ \hline 
Trains & 52.94\% & ~ 			& 5.436 \\ \hline 
Total & 72.55\% & 246.48 	& 14.994 \\ \hline 
~ & \multicolumn{3}{|c|}{Dictionary size 400}  \\ \hline 
~ & CQ & LT  & TT  \\ \hline 
Buses & 70.59\% & ~ & 4.116 	\\ \hline 
Cats & 88.24\% & ~ & 6.136 	\\ \hline 
Trains & 52.94\% & ~ & 5.344 	\\ \hline 
Total & 70.59\% & 265.469 & 15.596	\\ \hline 
\end{tabular}
\end{table}
In the BoF algorithm the learning process (dictionary generation) is run globally for all classes, thus column LT is empty in Table \ref{tab:BOF} for each class. 
In Table \ref{tab:results} we depict simulation results of the  method described in Sections \ref{sec:learning} and \ref{sec:testing}.
%
%
%
%
%
%
%
%
%
%
%
%
\begin{table}
\caption{Classification accuracy and time of the learning and testing processes obtained by the method proposed in this paper (CQ - Classification Quality ([\%]), LT - Learning time ([s]), TT - Testing time ([s])). The learning time is given only as the overall time for all classes.}
\label{tab:results}
\centering
\begin{tabular}{|p{0.7in}|p{0.6in}|p{0.6in}|p{0.6in}|p{0.7in}|p{0.7in}|} \hline 
\textbf{~} & Positive learning samples & Negative Learning samples & Classif. accuracy on testing set & Learning time ([s]) & Testing time: ([s]) \\ \hline 
Buses & 76 & 17 & 82.35\% & ~ & 3.236 \\ \hline 
Cats & 82 & 17 & 76.47\% & ~ & 4.495 \\ \hline 
Trains & 73 & 17 & 64.71\% & ~ & 3.593 \\ \hline 
Total & 231 & 51 & 74.51\% & 182.117 & 11.324 \\ \hline 
\end{tabular}
\end{table}
The BoF combined with the SVM algorithm achieved the best overall classification accuracy for three classes for the dictionary of size 350 (Table \ref{tab:BOF}), which was approximately worse by 2\%  than in the case of the proposed method. Moreover, the learning and classification time for the proposed method is considerably shorter than the BoF-SVM (182.117 s vs. 246.48 s), which is better, respectively by 35\% and 32\%.  The dictionaries of other sizes  performed slightly worse than the dictionary of size 350.  
It can be clearly seen that our method gives a better classification accuracy and the time of learning and testing process is shorter. It should be also emphasized that our method has an extra advantage; namely we can add a new class of visual objects to the existing system by just adding new rules. In the BoF we have to recreate the whole dictionary. The most time-consuming part of the bag-of-features classification is the SVM learning. 
\section{Conclusions}
We proposed a new approach to fast image classification. Our approach, which works by repeatedly creating fuzzy rules based on most salient image features, has shown promising results on a real-world dataset. Despite its simplicity, it outperformed the bag-of-features method in terms of accuracy and speed. It demonstrates the following advantages:
\begin{itemize}
\item the method is relatively accurate in terms of visual object classification,
\item learning and classification is very fast,
\item expanding the system knowledge is efficient as adding new visual classes to the system requires generation of new fuzzy rules whereas in the case of bag-of-features it requires new dictionary generation and re-learning of classifiers.
\end{itemize}
The method also demonstrates a potential in terms of possibility to expand it in order to incorporate different features or different meta learning algorithms. 

It should be noted that the system can work with virtually any type of fuzzy membership functions, e.g. triangular or bell-shape. Moreover,
various types of t-norm can be used in the algorithm but the application of the minimum t-norm leads to faster computation than in the case of other t-norms.

\section*{Acknowledgments}
This work was supported by the Polish National Science Centre (NCN)  under project number DEC-2011/01/D/ST6/06957.




\bibliography{CBIR_RS}
\bibliographystyle{elsarticle-harv}




\end{document}